# Backorder Prediction in Inventory Management: Classification Techniques and Cost Considerations

Sarit Maitra, Alliance Business School, Alliance University, Bengaluru, India, sarit.maitra@gmail.com
Sukanya Kundu, Alliance Business School, Alliance University, Bengaluru, India, sukanya.kundu@alliance.edu.in

**ABSTRACT**

This article introduces an advanced analytical approach for predicting backorders in inventory management. Backorder refers to an order that cannot be immediately fulfilled due to stock depletion. Multiple classification techniques, including Balanced Bagging Classifiers, Fuzzy Logic, Variational Autoencoder - Generative Adversarial Networks, and Multi-layer Perceptron classifiers, are assessed in this work using performance evaluation metrics such as ROC-AUC and PR-AUC. Moreover, this work incorporates a profit function and misclassification costs, considering the financial implications and costs associated with inventory management and backorder handling. The study suggests that a combination of modeling approaches, including ensemble techniques and VAE, can effectively address imbalanced datasets in inventory management, emphasizing interpretability and reducing false positives and false negatives. This research contributes to the advancement of predictive analytics and offers valuable insights for future investigations in backorder forecasting and inventory control optimization for decision-making.

**Keywords:** Backorder Forecasting; Cost Sensitive; Decision Science; Inventory Management; Machine Learning; Predictive Analytics;

## 1. INTRODUCTION

Backorders in inventory management refer to a customer's order for a product that is temporarily out of stock, resulting in a delay in fulfilment and delivery. Backorders can be both beneficial and detrimental. Orders might be delayed due to high demand, but so can poorly planning. Customers may not have the luxury or patience to wait if a product is not immediately available. This has a negative impact on sales and client satisfaction. Existing research (e.g., [1]; [2]) has experimented with machine learning (ML) to identify products at risk of backorders. However, there are still gaps that need to be addressed to improve the accuracy, timeliness, interpretability, and real-world implementation of backorder prediction models.

Companies are continually striving for a balance in the management of backorders. It's a tight line to walk: too much supply raises inventory expenses, while too little supply raises the danger of customers cancelling purchases. Why not maintain everything on hand at all times? Because most merchants and manufacturers have a significant number of SKUs (unique product IDs), this technique will increase inventory costs significantly.

The challenge here is to classify and forecast severely imbalanced product classes based on historical data from inventories and supply chains and to assess their propensity to have backorders. Despite the considerable research efforts ([1]; [2]; [3]; [4]) and researchers addressing backorder forecasting using artificial intelligence (AI) and machine learning (ML)-based prediction in inventory management, there are still open questions and challenges in this field. There is a lack of comprehensive studies that compare different AI-ML-based prediction systems in terms of their performance, advantages, limitations, and suitability for backorder forecasting. Moreover, while some studies look at financial consequences, the integration of cost and profit issues, such as misclassification costs and the profit function, is not thoroughly examined in the current literature. Many AI-based prediction models lack interpretability, making it difficult for decision-makers to comprehend the underlying issues driving backorders and take appropriate action. This work addresses these gaps by using highly imbalanced backorder data and advanced generative AI and machine learning (ML) techniques with the goal of bridging some of the existing gaps. The theoretical foundation of this article lies in the intersection of supply chain management, predictive analytics, and advanced AI-ML techniques, with a specific focus on backorder management and inventory system optimization.

This study considers several metrics, e.g., receiver operating characteristics area under the curve (ROCAUC), precision recall area under the curve (PRAUC), Macro F1-score (harmony between precision and recall), profit maximization, and misclassification cost, as cost-sensitive approaches to developing a final model. This comprehensive evaluation helps to assess the accuracy, robustness, and cost-effectiveness of the prediction models, offering important insights for decision-makers and practitioners in selecting the best applicable solutions for their specific supply chain contexts. The findings have important implications for organizations looking to improve their supply chain analytics capabilities to achieve operational excellence. Organizations can save costs, improve customer satisfaction through on-time delivery, mitigate the bullwhip impact, and enable proactive decision-making in changing market conditions by enhancing forecasting accuracy.

## 2. LITERATURE REVIEW

Before delving into demand and inventory management, we attempted to comprehend the key aspects influencing supply chain performance. Researchers discovered that supply chain structure, inventory management policy, information interchange, customer demand, forecasting method, lead time, and review period duration are important contributors in this context [5]. Our study incorporates all of these and eventually narrows down to a forecasting method that demonstrates interdependence with inventory management policy, customer demand, lead time, and review period distribution.

The stochastic demand has attracted attention from several scholars in the last decade. Their work encompasses various approaches such as parametric and non-parametric methods, artificial neural networks, FL-based techniques, and other diverse methodologies. Researchers used ML to increase the precision of backorder forecasts [6]. Their study demonstrated the value of ML, showing a 20% improvement in accuracy. The findings emphasized the flexibility, clarity, and enhanced precision that ML offers. Building on this work, few authors ([7], [8], [9], etc.) employed supervised ML techniques on the same dataset used in this research. Their study further corroborated the efficacy of ML in improving backorder prediction within SCM. They highlighted the need for continued research in exploring different algorithms, constructing a cost-sensitive learning framework, and validating performance enhancements. A recent study [8] explored the applications of AI and ML within supply chains which opens innovative avenues for further investigation on how AI and ML can be utilized in SCM.

The existing literature contains several works addressing the economic order quantity (EOQ) and economic production quantity (EPQ) models, considering the presence of backorders. For instance, a recent study conducted research on integrating the EOQ model with backorders and proposed an optimization algorithm to determine the optimal order quantity, backorder quantity, and reorder point [10]. Another study focused on the EPQ model with backorders and developed a mathematical model to optimize production quantity and backorder quantity [1]. All these studies contribute to backorder forecasting by integrating backorders into traditional inventory models, providing optimization techniques, offering practical applications, and leveraging advanced technologies. They enhance our understanding of managing backorders in inventory management, thereby enabling organizations to optimize their inventory systems, improve customer service levels, and make informed decisions.

In a different approach, a hybrid model was proposed that integrated ARIMA and ANN for backorder prediction [11]. By combining the strengths of both approaches, the hybrid model aims to improve the accuracy and robustness of backorder predictions. The impact of non-normal and autocorrelated demand on supply chain risk management was examined, and an empirical method was proposed for computing safety stock levels, showcasing improvements in cycle service level, inventory investment, and backorder volume [12]. They aim to provide a better estimation of safety stock levels, leading to improved cycle service levels, optimized inventory investment, and reduced backorder volume. A new Bayesian method was proposed based on compound Poisson distributions for demand forecasting that outperforms other methods [13]. Researchers also explored the use of deep learning techniques, specifically LSTM networks, for backorder prediction ([3], [12]). Their research highlighted the effectiveness of deep learning in capturing complex patterns and improving the accuracy of backorder forecasts. The combination of all the above methods demonstrates the advancement of backorder forecasting, contributing to improved forecasting.

Several authors (e.g., [14], [6], [1], [15], [3], [8]) have applied ML techniques to the same dataset to determine whether advanced ML techniques can improve the effectiveness of backorder forecasting in the early stages of the supply chain. Adaptable and resilient models are required to handle the complexity of massive inventory data and deliver the correct insights for decision-making ([16], [17]).

All these studies enhance our understanding of managing backorders, optimize inventory systems, improving customer service levels, and aiding decision-making. However, despite the progress made, the field of AI and ML in supply chain management, including backorder forecasting, is still in its early stages [18]. The challenges of working with big data in inventory management, such as data volume, variety, heterogeneity, and statistical biases, need to be addressed to develop adaptable and resilient models for accurate decision-making.

The literature review presented in this study serves as a foundation for developing the conceptual model displayed in Fig. 1 to further enhance backorder forecasting in supply chain management.

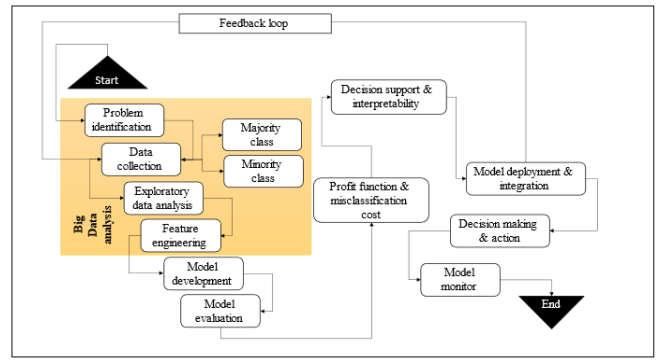

***Fig. 1***:  *Conceptual model*

Fig. 1 outlines a flow diagram with the different steps and measures were taken in our study to explore and validate the effectiveness of AI and ML techniques in backorder forecasting. By operationalizing the conceptual model into a logical framework, we aim to provide a structured framework for conducting empirical research, analyzing data, and deriving actionable insights for inventory management and supply chain decision-making. Furthermore, the logical model considers cost-sensitive learning frameworks, validation in real-world implementations, and considerations of contextual factors in backorder forecasting. It will also consider the financial implications and costs associated with inventory management and backorder handling by incorporating profit functions and misclassification costs into the model.

The problem statement can be hypothesized as follows: A hypothetical manufacturer has a data set that indicates whether a backorder has happened. The objective is to use predictive analytics and machine learning to reliably estimate future backorder risk and then discover the best method for inventorying products with high backorder risk.

## 3. DATA ANALYSIS & MODEL DEVELOPMENT

The inventory dataset used in this study has 1,04,8575 entries with 8 categorical and 15 numerical variables. A unique identification sku (stock-keeping unit) assigns each data point to a distinct product. SKUs assist businesses in precisely identifying and locating products, monitoring stock levels, and facilitating effective inventory management and order fulfilment operations. They are required for efficient inventory management, supply chain management, and sales analysis. This column, however, was eliminated because it provided no additional value for the purposes of this study. Table 1 shows a summary of the dataset's statistics. From the data type, we can identify the categorical columns (7 columns with object type, which excludes sku). The data file includes the past eight weeks' worth of data, which comes before the week we are attempting to forecast.

One challenge with the dataset is the imbalanced classes displayed in Fig. 2, where the majority class considerably outnumbers the minority class. We use SMOTE (synthetic minority over-sampling technique) and variational auto-encoder (VAE) to deal with unbalanced data sets, which increases modelling accuracy and efficiency. The second task is to optimize for the business case. To do this, we employ:

- Profit maximisation via classification models entails developing a profit function, optimizing the decision threshold, feature engineering, cost-sensitive learning, model selection, and continuous monitoring to ensure that the model's performance aligns with the business's financial objectives. It is a data-driven and business-focused strategy to increase profitability while considering any trade-offs and risks.
- We employ misclassification costs, which is a practical and business-oriented approach to optimization. It allows for informed decisions regarding model performance, model selection, and decision thresholds while considering the financial and operational aspects of the business case. The problem is viewed as a cross-sectional problem.

**Table 1:** Summary Statistics

| Sl no. | Variables | Description | Data type | Mean | Median | Std Dev | Maxima | Value |
|---|---|---|---|---|---|---|---|---|
| 2 | $nationalInv$ | Current inventory level | int64 | 489.42 | 15.00 | 28595.83 | 12334400 | unit |
| 3 | $leadTime$ | Transit time for product | float64 | 7.84 | 8.00 | 7.04 | 52 | weeks |
| 4 | $inTransitQty$ | Product in transit | int64 | 45.36 | 0.00 | 1390.53 | 489408 | |
| 5 | $forecast3Month$ | Forecast for next 3 months | int64 | 185.22 | 0.00 | 5032.30 | 1218328 | unit |
| 6 | $forecast6Month$ | Forecast for next 6 months | int64 | 360.88 | 0.00 | 10067.64 | 2461360 | unit |
| 7 | $forecast9Month$ | Forecast for next 9 months | int64 | 528.91 | 0.00 | 14895.45 | 3777304 | unit |
| 8 | $sales1Month$ | Sales revenue prior 1 month | int64 | 57.30 | 0.00 | 2067.93 | 741774 | unit |
| 9 | $sales3Month$ | Sales revenue prior 3 months | int64 | 180.46 | 0.00 | 5263.48 | 1094112 | unit |
| 10 | $sales6Month$ | Sales revenue prior 6 months | int64 | 352.46 | 0.00 | 9773.35 | 2146625 | unit |
| 11 | $sales9Month$ | Sales revenue prior 9 months | int64 | 544.33 | 0.00 | 15195.65 | 3201035 | unit |
| 12 | $minBank$ | Minimum amount to stock | int64 | 54.14 | 0.00 | 1244.24 | 313319 | unit |
| 13 | $potentialIssue$ | Past overdue | int64 | 3.28 | 0.00 | 299.43 | 146496 | unit |
| 14 | $piecesPastDue$ | Performance last 6 months | float64 | -7.05 | 0.82 | 26.84 | 1 | N/A |
| 15 | $perf6MonthAvg$ | Performance last 12 months | float64 | -6.62 | 0.80 | 26.14 | 1 | N/A |

| 16 | *perf12MonthAvg* | Amount of stock orders overdue | int64 | 0.63 | 0.00 | 35,18 | 12530 | unit |
| 17 | *localBoQty* | Issue identified | object | | | | | |
| 18 | *deckRisk* | Risk flag | object | | | | | |
| 19 | *oeConstraint* | Risk flag | object | | | | | |
| 20 | *ppapRisk* | Risk flag | object | | | | | |
| 21 | *stopAutoBuy* | Risk flag | object | | | | | |
| 22 | revAtop | Risk flag | object | | | | | |
| 23 | wentOnBackorder | Product went on backorder | object | | | | | |

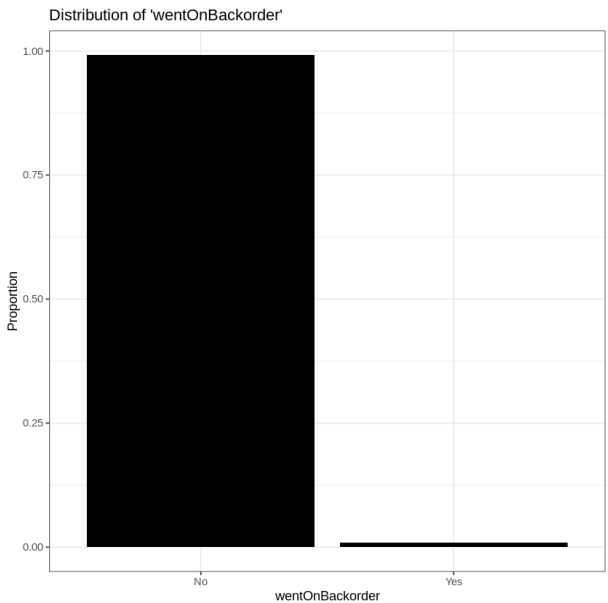

*Fig. 2: Severe imbalanced classes*

### 3.1 Exploratory Data Analysis

- A thorough data analysis was done for this dataset, which helped in understanding the characteristics, patterns, and relationships within the data.

- The attribute indicates the current inventory levels. The mean inventory is 489.42 units, with a wide range from 15 units to a maximum of 12,334,400 units. The median value is 15 units, indicating a possible right-skewed distribution.

- The attribute represents the transit time for the product. The mean transit time is 7.84 weeks, with a median of 8 weeks. The standard deviation is 7.04 weeks, indicating some variability in transit times.

- The attribute indicates the quantity of products currently in transit. The high standard deviation and maximum value indicate significant variation in transit quantities.

- The dataset includes variables representing sales revenue for different time periods (1 month, 3 months, 6 months, and 9 months) as well as forecasted sales for the next 3, 6, and 9 months.

- The mean and median values indicate the average and middle values of sales and forecasts. The high standard deviations suggest variability in sales performance.

- The $minBank$ attribute represents the minimum amount required to stock. The mean and median values indicate the average minimum stocking requirement, while the high maximum value suggests that some products may require a significantly higher minimum stock level.

- The $piecesPastDue$ attribute represents the quantity of products that are past-due. The mean and median values suggest a low average quantity of past due items.

- The $perf6MonthAvg$ and $perf12Monthavg$ attributes represent the performance of products over the last 6 and 12 months, respectively. The negative mean values for both variables indicate below-average performance, while the standard deviations suggest variation in performance.

- The $localBoQty$ attribute indicates the number of stock orders overdue. The low mean and median values suggest a relatively low average amount of overdue stock orders.

- Several attributes, such as $potentialIssue$, $deckRisk$, $oeConstraint$, $ppapRisk$, $stopAutoBuy$, and r$evStop$, represent risk flags associated with the products. These variables are categorical and indicate the presence or absence of specific risks.

- The $wentOnBackorder$ attribute represents whether a product went on backorder. This variable is the target variable, and further analysis is required to understand its distribution and relationship with other variables.

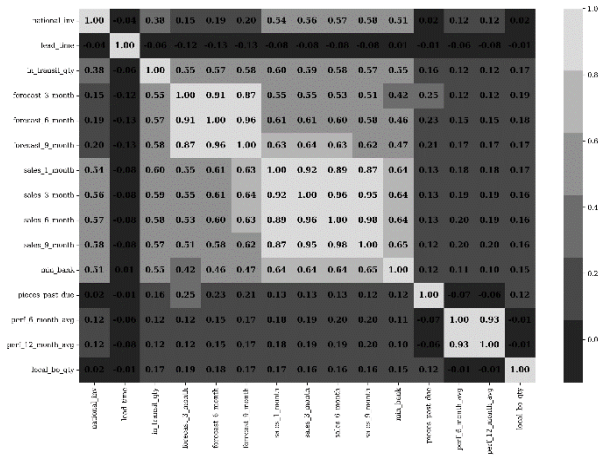

*Fig. 3: Correlation heatmap of numerical features*

Fig. 3 displays the Spearman correlation matrix of numerical attributes. All the significant correlations observed are positive.

- $forecast3Month$, $forecast6Month$ and $forecast9Month$ are strongly correlated (coefficient = 0.99).
- $sales1Month$, $sales3Month$, $sales6Month$ and $sales9Month$ are strongly correlated with each other with a degree varying from 0.82 to 0.98.
- forecast and sale columns are correlated with each other with a minimum degree of 0.62 varying up to 0.88. It is obvious that when the sales for a certain product are high in the past sales the forecast for the same in the coming months will be higher and vice versa.
- $perf6MonthAvg$ and $perf12MonthAvg$ are very highly correlated with each other (coefficient = 0.97.
- $minBank$ (minimum amount of stock recommended) is highly correlated with sales and forecast columns as stock in inventory is directly proportional to sales.
- $inTransitQty$ is highly correlated with sales, forecast and $minBank$ columns. This is obvious because high sales of a product => more of that product in transport for inventory replenishing high sales of a product => high forecast.
- $piecesPastDue$ is meekly correlated with sales and forecast columns $nationalInv$ is meekly correlated with $minBank$ and weekly correlated with sale columns.

Overall, the correlation matrix indicates that the number of features used to forecast whether an item will be placed in backorder may be less than the number of features in the data set. Fig. 4 displays the correlation matrix of categorical attributes.

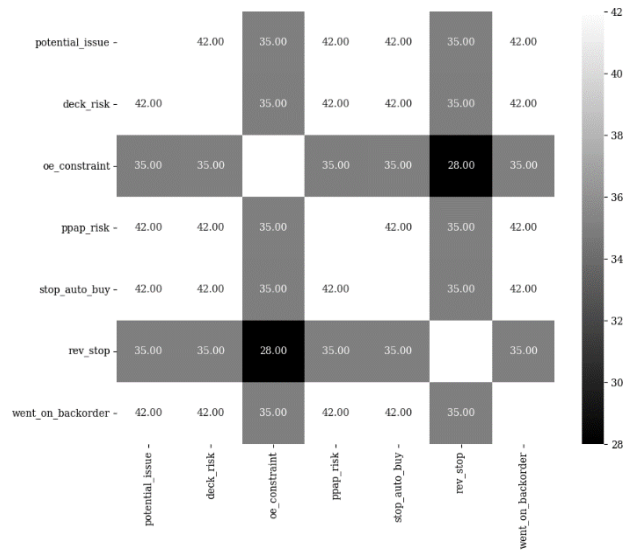

*Fig. 4: Chi-Squared test heatmap of categorical features*

Based on the chi-squared values, there is no strong association between any pairs of variables. There might be a relationship between $oeConstraint$ and $potentialIssue$. $RevStop$ and $oeConstraint$ may be related in some way. There might be a relationship between $potentialIssue$ and $revStop$. Even if the coefficients are high, the features mentioned above have the lowest scores in comparison.

Several data pre-processing steps were employed to get the raw data ready for modelling. This includes replacing -99.0 in performance columns with nan for imputing and employing an iterative imputer to fill in the missing values. Moreover, as some of the features in the numerical columns are correlated, linear models like Logistic Regression, Linear SVM (support vector machines), and other linear models may not perform as well as the coefficients of separating plane change.

The values in the category columns are one hot encoded with "Yes" as 1 and "No" as 0. There are missing values in the columns $leadTime$, $perf6MonthAvg$, and $perf12MonthAvg$. To fill in the missing variables, model-based imputation (iterative-imputer) was employed. Moreover, the positive skewness (possible outliers) in the data was treated in two ways, resulting in two distinct datasets to apply.

- Values = (value - median) / (75% value - 25% value) to standardize the data while scaling it without taking outliers into account.
- Applied the log transformation followed by the Standard Scaler to the columns in the dataset with positive skewness.

Principal Component Analysis (PCA) was employed to identify major features explaining 99% variance in the dataset. Fig. 5 displays the top three features in our PCA are: $nationalInv$, $forecast9Month$ and $sales9Month$.

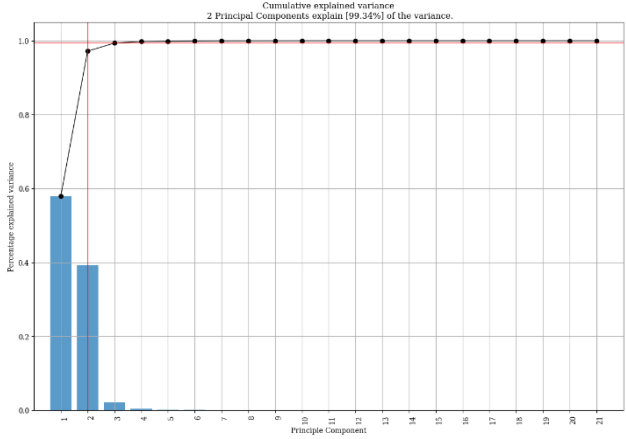

***Fig. 5:*** *Feature importance using PCA.*

The performance metrics used in this work are: ROCAUC, which differentiates between positive and negative classes. PRAUC, which helps in selecting an appropriate threshold that balances the trade-off between precision and recall. The Macro F1-Score is the average of the F1 scores of both the positive and negative classes.

Besides these three primary metrics, our work reports Mathews Correlation Coefficient (MCC), precision, recall, and Brier scores for better comparison. MCC considers true positive (TP), true negative (TN), false positive (FP), and false negative (FN) values to provide a balanced measure of classification accuracy. Considering these, our work aims to provide a comprehensive evaluation of the predictive models in terms of their performance, interpretability, and latency constraints. By not having latency constraints, the models can potentially make more accurate and informed predictions by leveraging a broader range of data and capturing any patterns or trends that may emerge over longer time periods. It allows for a more comprehensive analysis of the backorder situation and better forecasting of the likelihood of products going into backorder.

### 3.2 Hypotheses test

We employed non-parametric Mann-Whitney and Chi-Square tests to help assess the association between each attribute and the target variable, or their differences between different groups. The results are displayed in Table 2.

The results indicate significant associations between certain attributes and backorders. Low chi-square test p-values for the attributes $deck\_risk$, $oe\_constraint$, and $ppapRisk$ indicated strong relationships.

All the numerical attributes displayed significant relationships with backorders, based on the Mann-Whitney U tests. These findings suggest that these attributes may play a crucial role in predicting backorders.

### 3.3 Summary of data analysis and feature engineering

We have a severely imbalanced dataset with 99.99% majority class. The problem we are trying to address is a binary classification problem where we have to predict whether or not a product will go to backorder.

***Table 2:*** *Statistical hypotheses tests*

| Sl. no. | Attributes | Statics with Significance |
|---|---|---|
| 0 | $nationalInv$ | U statistic: 1834036383.5, p-value: 0.00 |
| 1 | $leadTime$ | U statistic: 4239446959.0, p-value: 0.00 |
| 2 | $inTransitQty$ | U statistic: 7002852514.5, p-value: 0.00 |
| 3 | $forecast3Month$ | U statistic: 6905403097.0, p-value: 0.00 |
| 4 | $forecast6Month$ | U statistic: 6827462574.5, p-value: 0.00 |
| 5 | $forecast9Month$ | U statistic: 5869355615.0, p-value: 0.00 |
| 6 | $sales1Month$ | U statistic: 5910651312.5, p-value: 0.00 |
| 7 | $sales3Month$ | U statistic: 5809789939.0, p-value: 0.00 |
| 8 | $sales6Month$ | U statistic: 5740565459.0, p-value: 0.00 |
| 9 | $sales9Month$ | U statistic: 4688282621.5, p-value: 0.019 |
| 10 | $minBank$ | U statistic: 5062287539.5, p-value: 0.00 |
| 11 | $potentialIssue$ | U statistic: 4193563375.0, p-value: 0.00 |
| 12 | $piecesPastDue$ | U statistic: 4174380247.5, P-value: 0.00 |
| 13 | $perf6MonthAvg$ | U statistic: 5145229552.0, P-value: 0.00 |
| 14 | $perf12MonthAv$ | U statistic: 1834036383.5, p-value: 0.00 |
| 15 | $localBoQty$ | U statistic: 4239446959.0, p-value: 0.00 |
| 16 | $deckRisk$ | Test statistic: 219.903, p-value: 0.00 |
| 17 | $oeConstraint$ | Test statistic: 28.274, p-value: 0.00 |
| 18 | $ppapRisk$ | Test statistic: 110.458, p-value: 0.00 |
| 19 | $stopAutoBuy$ | Test statistic: 4.564, p-value: 0.032 |
| 20 | $revAtop$ | Test statistic: 2.978, p-value: 0.084 |

The Dataset contains 15 numerical attributes all of which are highly skewed.
- $leadTime$ attributes come with missing values (64518).
- $perf6MonthAvg$ and $perf12MonthAvg$ are heavily left skewed rest all the features are right skewed.

- The numerical attributes have a small interquartile range, and some have negative values.
- The attributes of sales, forecast and performance are correlated.
- Dataset contains 8 categorical attributes, including *wentToBackorder* is the target variable.
- Encoded categorical attributes to numerical attributes.
- Dataset split into train and test dataset with 80:20.
- Iterative imputer used for missing value imputation.
- Performed PCA to determine feature importance.
- Different feature transformation was applied on the dataset.
  - Robust scaler
  - Log transform
  - Standard scaler

### 3.4 Modelling and Evaluation

We have experimented with multiple models (5-models) to address the complexities and challenges of inventory management systems. There is no one size fits for all and scholars have experiments with different techniques in their work. Such as distributed Random Forest and Gradient Boosting [6], deep neural network was also proposed [3], The models used are the dummy model as the base model, the Balanced Bagging classifier (BBC), the BBC with variational auto-encoder, Fuzzy logic with the BBC, Random Forest, and Multilayer Perceptron. Each model was implemented with four transformed datasets (Robust Scaling, log Transform and Standard Scaling, Quantile Transform). Thus, a total of 20 models were trained and tested.

To start with, a grid search cross-validation was employed to fine-tune the Balanced Bagging Classifier (BBC) by exploring different hyperparameters. We used Grid Search Cross-Validation (GridSearchCV) for optimizing hyperparameter tuning by testing alternative hyperparameter combinations. It combines grid search, which specifies a range of hyperparameter values, with cross-validation, which evaluates the model's generalization to previously unknown data. The optimum hyperparameters are chosen depending on the ROC AUC. Because of its scientific approach and avoidance of human biases, GridSearchCV is a popular method for hyperparameter tuning among the scientific community and applied fields. We performed an exhaustive grid search using all the combinations of parameters to optimize the internal score in the training set. Table 3 displays the pseudocode for hyperparameter search.

*Table 3: Pseudocode - exhaustive hyperparameter search*

```
# Hyperparameters and the possible values
parameters = {
    'n_estimators': [20, 50, 100, 200, 300, 400, 500, 600, 700, 800, 900, 1000, 1200],
    'max_features': [0.1, 0.2, 0.3, 0.4, 0.5, 0.6, 0.7, 0.8, 0.85, 0.90, 0.92, 0.95, 1.0],
    'bootstrap': [True, False],
    'bootstrap_features': [True, False],
}

# BBC Instance
classifier = Balanced Bagging Classifier ()

# Grid Search cross validation instance
gridSearch = GridSearchCV(estimator = classifier, param_grid = parameters, cross validation = 5, scoring = 'roc_auc')

# Fit GridSearchCV to the training data
gridSearch.fit(X_train, y_train)

# Visualize the results
gridSearchPlot(grid_search)
# Function to create grid search plots
gridSearchTable(grid_search)
# Function to create a summary table of results
```

The purpose of controlling the maximum number of features during grid search is to introduce randomness and diversity into the individual base estimators of the ensemble. It helps to prevent overfitting and improve the generalization of the ensemble. The idea is that by restricting the number of features considered at each split, the base estimators are forced to make more independent decisions, which can lead to a more robust and accurate ensemble model. Fig. 6 displays the model after the grid search parameters. Here, the ensemble consists of 1000 trees, and n_jobs, which controls parallel processing, is set to -1 to use all available CPU cores for parallel execution.

```
BalancedBaggingClassifier
BalancedBaggingClassifier(n_estimators=1000, n_jobs=-1)
```

*Fig. 6: Balanced bagging classifier*

BBC is an ensemble learning method that combines multiple classifiers, which increases the performance and generalization of the model. The BBC combines the advantages of bagging and sampling techniques to address the issue of imbalanced datasets. This has been supported by researchers in the past [19].

Several researchers in recent times and in the past have recommended fuzzy logic (e.g., [11], [21], [22], [23], [24], [25], [26]). The objective is to add human-centric design along with advanced machine learning algorithms. When dealing with imbalanced data or when the problem exhibits complicated and ambiguous linkages, employing fuzzy logic in conjunction with balanced bagging might be a beneficial method. Therefore, fuzzy logic was integrated into the BBC to efficiently handle imprecise information that is frequently encountered in inventory management systems.

Furthermore, we leveraged the power of the Generative Adversarial Network—Variational Auto Encoder (VAE)—to experiment with our modelling approach. Researchers claimed the superior performance of VAE compared to supervised ML techniques for imbalanced data (e.g., [27], [28]). Their work recommends VAE, offering great potential for mitigating the bullwhip effect and enhancing supply chain operations. We combined supervised BBC with unsupervised VAE to develop a powerful approach for addressing class imbalance and capturing complex feature representations. VAE generates meaningful latent representations of the input data, which are then combined with the original features to improve the performance of the BBC. Table 4 displays the pseudocode for VAE implementation.

*Table 4: Pseudocode for handling imbalanced data with VAE and BBC model training*

```
# Data Splitting
X, y = SplitData(df);

# Data Preprocessing
numeric_features = ['nationalInv', 'leadTime',
'inTransitQty', 'piecesPastDue', 'localBoQty'];
categorical_features = ['potentialIssue', 'deckRisk',
'oeConstraint', 'ppapRisk', 'revStop'];

# VAE model
input_dim = GetNumberOfFeatures(X);
latent_dim = 10;

# Encoder and decoder layers for the VAE
"""Encoder and decoder layers of the VAE are defined
with activation functions. These layers are part of the
neural network architecture used in the VAE"""
input_layer = DefineInputLayer(input_dim);
encoded =
DefineDenseLayerWithActivation(input_layer, 32,
'relu');
encoded = DefineDenseLayerWithActivation(encoded,
latent_dim, 'relu');
decoded = DefineDenseLayerWithActivation(encoded,
32, 'relu');
decoded = DefineDenseLayerWithActivation(decoded,
input_dim, 'linear');

# VAE model with custom loss
vae = CreateVAEModel(input_layer, decoded,
CustomLossFunction);

# Train VAE on the imbalanced data
TrainVAEModel(vae, X_train_processed,
X_train_processed, epochs=20, batch_size=32);

# Encode the input data using the VAE
encoded_X_train = EncodeDataWithVAE(encoder,
X_train_processed);

# Combine the latent representations with the original
features
combined_X_train =
ConcatenateFeatures(X_train_processed,
encoded_X_train);

# Train BBC on the combined feature set
bbc =
CreateBalancedBaggingClassifier(n_estimators=1000)
TrainBalancedBaggingClassifier(bbc,
combined_X_train, y_train);

# Preprocess the test data and obtain predictions
X_test_processed = PreprocessTestData(X_test);
encoded_X_test = EncodeDataWithVAE(encoder,
X_test_processed);
combined_X_test =
ConcatenateFeatures(X_test_processed,
encoded_X_test);
y_pred = PredictWithBalancedBaggingClassifier(bbc,
combined_X_test);

# Calculate ROC AUC score
roc_auc = CalculateROCAUCScore(y_test, y_pred);
Print("ROC AUC score:", roc_auc);
```

Lastly, we tried Artificial Neural Network based Multilayer Perceptron (MLP). MLP provides efficient computing, lowering computational time and memory utilization, making it a desirable tool for real-world inventory control systems ([13]; [28]).

## 4. RESULTS & DISCUSSIONS

Table 5 presents a consolidated report of all the classifier experimented.

- *ROCAUC*

    a) *on non-normal data*: BBC (0.9081) > VAE_BBC (0.9003) > FL_BBC (0.8759) > Dummy (0.5074).

    b) *on Log-transformed and Normalized data:* BBC (0.9073) > VAE_BBC (0.9007) > FL_BBC (0.8615) > Dummy (0.4897).

- *PRAUC*

    a) *on non-normal data*: BBC (0.4925) > VAE_BBC (0.4841) > FL_BBC (0.4646) > Dummy (0.2640).

    b) *PRAUC on Log-transformed and Normalized data:* BB (0.4917) > VAE_BBC (0.4847) > FL_BBC (0.4515) > Dummy (0.2456).

- *Macro F1-Score*

    a) *on non-normal data*: BB (0.5545) > VAE_BBC (0.5532) > FL_BBC (0.5251) > Dummy (0.3407).

    b) *on Log-transformed and Normalized data:* BB (0.5544) > VAE_BBC (0.5524) > FL_BBC (0.5195) > Dummy (0.0160).

- *Precision*

- a) *on non-normal data*: BB (0.0838) > VAE_BBC (0.0824) > FL_BBC (0.0601) > Dummy (0.0087).
- b) *on Log-transformed and Normalized data: BB (0.0840) > VAE_BBC (0.0818) > FL_BBC (0.0555) > Dummy (0.0081)*.

- *Recall*
  - a) *on non-normal data*: BBC (0.9005) > VAE_BBC (0.8848) > FL_BBC (0.8679) > Dummy (0.5151).
  - b) *on Log-transformed and Normalized data*: BB (0.9017) > VAE_BBC (0.8865) > FL_BBC (0.8460) > Dummy (0.4786).

- *Mathew's correlation coefficient*
  - a) *on non-normal data*: BB (0.2600) > VAE_BBC (0.2552) > FL_BBC (0.2099) > Dummy (0.0027).
  - b) *on Log-transformed and Normalized data*: BB (0.2609) > VAE_BBC (0.2544) > FL_BBC (0.1976) > Dummy (-0.0037).

- *Brier score*
  - a) *on non-normal data*: Dummy (0.2500) > VAE_BBC (0.0626) > BB (0.0623) > FL_BBC (0.0815).
  - b) *on Log-transformed and Normalized data*: Dummy (0.2500) > VAE_BBC (0.0629) > BB (0.0622) > FL_BBC (0.1310).

*Table 5: Results & discussions*

| | ROCAUC | PRAUC | Macro F1-Score | Precision | Recall | MCC | Brier |
|---|---|---|---|---|---|---|---|
| **Retained original dimensions** | | | | | | | |
| *Models on non-normal data* | | | | | | | |
| BBC | 0.9081 | 0.4925 | 0.5545 | 0.0838 | 0.9005 | 0.2600 | 0.0623 |
| FL_BBC | 0.8759 | 0.4646 | 0.5251 | 0.0601 | 0.8679 | 0.2099 | 0.0815 |
| VAE_BBC | 0.9003 | 0.4841 | 0.5532 | 0.0824 | 0.8848 | 0.2552 | 0.0626 |
| *Models on Log transformed + Normalized dataset* | | | | | | | |
| Dummy | 0.4897 | 0.2456 | 0.0160 | 0.0081 | 0.4786 | -0.0037 | 0.2500 |
| BBC | 0.9073 | 0.4917 | 0.5544 | 0.0840 | 0.9017 | 0.2609 | 0.0622 |
| FL_BBC | 0.8615 | 0.4515 | 0.5195 | 0.0555 | 0.8460 | 0.1976 | 0.1310 |
| VAE_BBC | 0.9007 | 0.4847 | 0.5524 | 0.0818 | 0.8865 | 0.2544 | 0.0629 |
| Random undersampler with | 0.9604 | 0.2428 | 0.5483 | 0.0785 | 0.9005 | 0.2507 | 0.0641 |

For a severely imbalanced dataset, the benchmark metrics most used are Area Under the Precision-Recall Curve (PRAUC), F1-Score, Matthews Correlation Coefficient (MCC), ROC AUC, Precision at a Given Recall, and Geometric Mean ([29], [30]). Here we have used a combination of multiple metrics, which provides a more comprehensive assessment of model performance.

Based on the evaluation metrics, it appears that the BBC model on the log-transformed and normalized dataset performed well across multiple metrics, indicating its effectiveness in predicting the target variable. Additionally, the "VAE + BBC" model on non-normal data also showed promising performance. In most cases, a higher value indicates better prediction performance, except for the Brier score, where a lower value is preferred. The Brier score measures the accuracy of probabilistic predictions, and a lower score indicates better calibration and accuracy.

Figs. 1 and 2 provide valuable insights into the performance of the models. The confusion matrix (CM) in Figure 1 allows us to examine the model's predictions in detail. It shows the number of true negatives (TN) and false positives (FP) for backorder predictions.

In this case, the high number of TNs (190,429) indicates that the model performs well in accurately identifying non-backorders. This means that the model correctly identifies a significant portion of the instances where a backorder is not present, which is crucial for efficient inventory management.

The number of FP (17,506), however, indicates that there are some situations in which the model forecasts non-backorders as backorders. This can lead to unnecessary actions or costs associated with backorder handling for those specific instances.

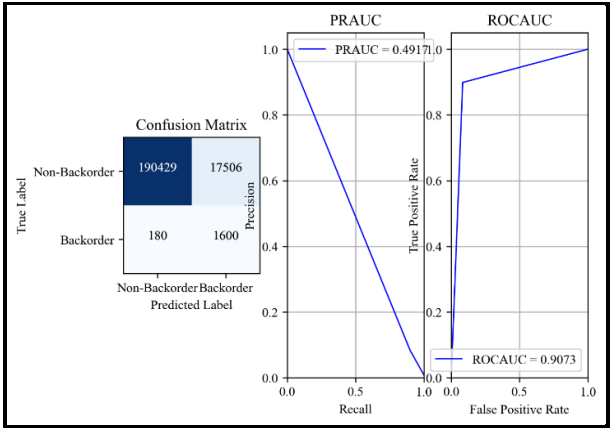

***Fig. 7:*** *VAE + BBC model - confusion matric and roc curves*

The presence of false negatives (180) indicates the model occasionally fails to predict backorders when they occur. This means that there are instances where the model incorrectly classifies a backorder as a non-backorder. This can result in missed opportunities to take proactive measures and prevent backorders, leading to potential disruptions in the supply chain and customer dissatisfaction. On the other hand, the presence of true positives (1600) suggests that the model can identify backorders to some extent. These instances represent the correct classification of backorders by the model.

Conversely, Fig. 2 shows that a high number of TN indicates that the model properly recognised a large number of examples (190,235) as non-backorders, indicating that it works well in reliably identifying non-backorders. A moderate number of FP means the model misclassified a considerable number of instances as backorders when they were non-backorders (17,700). This suggests that the model tends to predict false positives, which means it may sometimes predict an item as a backorder when it is not.

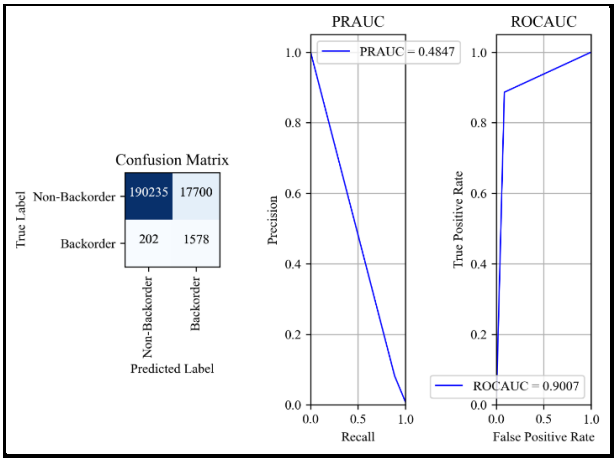

***Fig. 8:*** *BBC-A (non-normal data)*

A low number of FPs means the model misclassified a relatively small number of instances as non-backorders when they were backorders (202).

A moderate number of TPs means the model correctly identified a reasonable number of instances as backorders (1,578). This indicates that the model can predict backorders accurately. Overall, the model shows good performance in correctly identifying non-backorders (high TN) but has some limitations in accurately predicting backorders (moderate TP and FN). Reducing false positives and false negatives would be valuable for improving the overall accuracy of backorder predictions.

Matrix 2 performs slightly better than Matrix 1, as it has a lower number of FP and FN and a slightly higher number of TP. However, the differences between the two matrices are relatively small. This means that for Matrix 1, around 88.64% of products identified as potential backorders went on backorder, while for Matrix 2, around 90.02% of products identified as potential backorders went on backorder.

In this work, we justify using supervised learning under the presumption that the memory that is accessible is limited in relation to the size of the dataset. The contexts of big data, distributed databases, and embedded systems all apply to this fundamental paradigm. We study a simple yet powerful ensemble framework in which each individual ensemble model is constructed from a random patch of data generated by selecting random subsets of features and instances from the entire dataset. These results demonstrate that the suggested method performs on par with popular ensemble methods in terms of accuracy while simultaneously lowering memory requirements and achieving much superior performance when memory is severely limited.

### 4.1 Profit function

By measuring the overall profit generated by the forecasts, we can gain insights into the financial impact of the model's predictions and make informed decisions about inventory levels, ordering quantities, and backorder management strategies. This evaluation goes beyond traditional accuracy metrics and provides a more comprehensive assessment of the model's performance from a business perspective. The profit calculation in the given formula involves the following parameters: $SalesRevenue\ (1, 3, 6, and\ 9\ months)$, $holdingCost$, $inventoryLevel$, $backorderCost$, $leadTimeCost$, $potentialIssueCost$, $deckRiskCost$.

$$Revenue = \sum (sales1Month + sales3Month + sales6Month + sales9Month) \quad (1)$$

$$Profit = Revenue - (holdingCost * inventoryLevel) - (backorderLost * backorders) - (leadTimeCost * leadTime) - (potentialIssueCost * potentialIssue) - (deckRiskCost * deckRisk) \quad (2)$$

We have performed optimization to maximize the profit by finding the optimal values for the decision variables: $holdingCost$, $backorderCost$, $leadTimeCost$, $potentialIssueCost$, and $deckRiskCost$. The initial guess for the decision variable values was set to $x_0$. Here, we have defined a constraint as $holdingCost \geq 0$. The goal is to find the optimal values for the decision variables (costs) that maximize the profit. Results are displayed in Table 6. We have used numerical optimization method by employing Sequential Least Squares Quadratic Programming (SLSQP).

**4.2 Cost Sensitive Learning**

The overall cost of misclassification computed as the sum of the costs of false positives and false negatives. It is calculating the entire cost of inaccurate predictions, taking into consideration the expenses associated with each form of misclassification.

- $FPcost = 10$ (cost of misclassifying a backorder item as non-backorder)
- $FNcost = 1$ (cost of misclassifying a non-backorder item as backorder)

$$Total\ Cost = \{FP * FPcost\} + \{FN * FNcost\} \quad (3)$$

*Table 6: Cost perspective*

| Classifier | Optimal Profit | Misclassification cost |
|---|---|---|
| *Non-normal data* | | |
| BBC-A | 98,37,422.94 | 1,75,443.00 |
| FuzzyBBC-A | 74,51,418.43 | 2,42,238.00 |
| VAE + BBC | 98,37,424.40 | 1,78,353.00 |
| *Log normalized data* | | |
| BBC-B | 98,37,510.94 | 1,75,590.00 |
| FuzzyBBC-B | 74,51,422.38 | 2,42,233.00 |
| VAE + BBC | 98,37,423.06 | 1,78,060.00 |
| MLP | 98,36,319.84 | 1,80,041.00 |

Table 4 displays the cost sensitive factors. Considering both average profit and misclassification cost, the VAE + BBC model appears to perform well in terms of optimal profit, while the BBC-A model performs well in terms of misclassification cost for non-normal data.

The VAE model consists of two main components: an encoder and a decoder. The encoder takes the input data and transforms it into a lower-dimensional latent space representation, capturing the underlying patterns and features. The encoder converts the input data $x$ to a latent space representation $z$ that is parameterized by $mean$ () and variance (2):

$$q\ (z \mid x) = N\ (z \mid (x), 2(x)) \quad (4)$$

This latent space representation is then used as input for the decoder. Based on the learned latent representations, the decoder provides a meaningful and accurate reconstruction of the original data. This reconstruction method aids in capturing the key aspects and qualities of the data, allowing the model to generate correct predictions and classify instances. $p\ (x \mid z) = N\ (x \mid f(z))$, where $f(z)$ denotes the mapping from the latent space to the original feature space. The VAE aims to maximize the ELBO objective, which is a trade-off between reconstruction accuracy and regularization of the latent space:

$$ELBO = E\ [log\ p\ (x \mid z)] - D\_KL(q(z \mid x) \mid\mid p(z)) \quad (5)$$

where $E[log\ p(x \mid z)]$ represents the reconstruction term and $D\_KL(q(z \mid x) \mid\mid p(z))$ is the KL divergence between the approximate posterior $q(z \mid x)$ and the prior distribution $p(z)$.

BBC is further integrated with the decoder to improve the overall performance. Assuming a set of base classifiers as $C = \{C1, C2, \ldots, CM\}$, where M is the number of base classifiers. Given an input sample $x$, each base classifier $C_i$ provides a class prediction $C_i(x)$. The ensemble prediction is determined by majority voting, where the class label with the highest number of votes is selected:

$$ensemble\ prediction = argmax\ (class\_label, \sum[Ci(x) == class\_label]) \quad (6)$$

This approach ensures that the final prediction considers the opinions of multiple classifiers. It is evident that the combination of these techniques resulted in superior performance for our use case.

**4.3 Interpretability**

Interpretability is a critical aspect of the adoption of AI and ML in business ([31], [32]). It enables supply chain professionals to choose the best ML algorithms and comprehend the justification for the forecasts and decisions made by the model. This understanding empowers practitioners to optimize supply chain operations, improve decision-making, enhance visibility, mitigate risks, and drive overall efficiency and effectiveness ([33], [34]). By embracing interpretability, practitioners can make informed adjustments, ensure regulatory compliance, and establish trust with stakeholders, thus harnessing the full potential of AI and ML in transforming supply chain management.

We employed permutation importance (PI) to understand the relative importance of each attribute while making decisions. PI measures the importance of each feature by randomly shuffling its values and observing the impact on the model's performance This helps to understand the attributes affecting backorder predictions and sheds light on the contribution of each feature in the model. Fig. 7 displays the relative weights of each attribute in predicting backorders. These weights, along with their standard deviations, provide insights into how each attribute influences the model's output. We can draw several key conclusions from this analysis:

- $nationalInv$ attribute carries a high weight, indicating its significant impact on the model's predictions. Changes in this attribute (national inventory) values have a substantial effect on anticipating backorders. This underscores the pivotal role that inventory levels play in predicting backorders.
- $sales1Month$ is the second most influential attribute is "sales1Month." Its weight suggests that variations in sales over a one-month period significantly affect the model's predictions. The speed of recent sales activity appears to be a critical factor.
- $forecast3Month$ and $Sales9Month$ attributes also exhibit substantial importance, emphasizing the value of both short-term (3-month) forecasts and the history of sales over a 9-month period in predicting backorders.

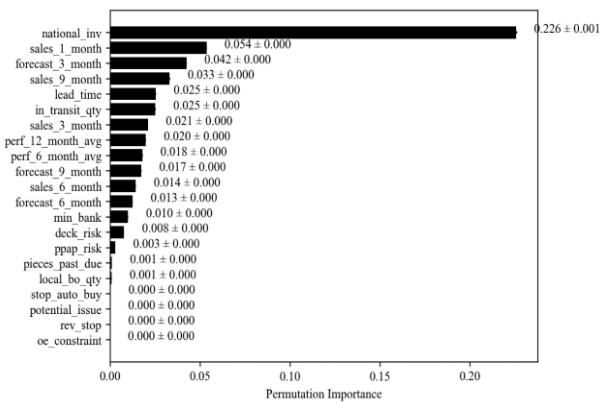

*Fig. 9: Permutation Importance (BBC model)*

$nationalInv$ is a critical attribute because it reflects a company's readiness to fulfil customer demand without encountering backorders. Maintaining optimal inventory levels is essential for operational efficiency and customer satisfaction. The high weight assigned to $nationalInv$ in the model's predictions underscores its significance in the context of backorder prediction, as it directly impacts a company's ability to meet customer demand and avoid disruptions in its operations. Likewise, $sales1Month$ attribute captures recent demand patterns and market dynamics, allowing the model to make informed predictions about backorders. Its significance lies in its ability to reflect immediate sales trends and enable timely responses to changes in customer demand, ultimately contributing to customer satisfaction and operational efficiency. $forecast3Month$ attribute provides forward-looking insights into expected demand, aiding in inventory planning and procurement decisions. Meanwhile, $Sales9Month$ attribute offers a long-term historical context, enabling the model to assess stability, analyze cyclical patterns, and build resilience in the face of changing market conditions. Both attributes contribute significantly to the accuracy of backorder predictions by considering various aspects of demand and sales behavior.

Some of the features, such as $StopAutoBuy$, $PpapRisk$, $revStop$, and $OeConstraint$, carry nearly zero weights. This indicates that changes in these attributes have minimal impact on backorder predictions and are less significant in this context.

### 4.4 Limitations

The study relies heavily on data quality and availability, which could be improved by overcoming data-related challenges. The real-world supply chain dynamics are complex and ever-changing, and future research could explore adaptive models that can adapt to dynamic environments in real-time. The findings are promising, but their applicability across diverse industries and supply chain structures may vary. Future research directions include developing explainable AI models, real-time supply chain optimization, hybrid models, cost modelling, and cross-domain applications.

The study paves the way for a new era of supply chain management by integrating state-of-the-art technologies and financial considerations. We encourage researchers and practitioners to embrace these limitations as opportunities for further innovation. The future of supply chain analytics is bright, with groundbreaking discoveries and transformative applications.

## 5 Conclusions

This paper presents a novel perspective on the impact of advanced predictive analytics techniques on supply chain management performance. By combining generative AI-based unsupervised VAE with the supervised BBC model, the study offers valuable insights into proactive strategies, stockout mitigation, and supply chain optimization. The research demonstrates the trade-off between profit and misclassification costs, providing a comprehensive evaluation of the model's performance through various metrics such as ROCAUC, PRAUC, Macro F1-Score, Precision, Recall, MCC, and Brier. By incorporating a profit function and considering misclassification costs along with these metrics, this work addresses the financial implications and costs associated with inventory management and backorder handling. Additionally, the inclusion of permutation importance enhances the interpretability of the model by identifying the relative importance of input features. This provides valuable insights into the factors influencing backorder forecasting and inventory control, contributing to future investigations in the field. Looking ahead, the potential real-world applications of our research are vast. Businesses operating in dynamic supply chain environments can leverage our insights to fine-tune their inventory management strategies, minimize stockouts, and enhance customer satisfaction. Moreover, the findings have the potential to revolutionize the way business make decisions by incorporating cost-sensitive predictive analytics into their daily operations.